\documentclass{article}

\usepackage{PRIMEarxiv}

\usepackage{enumitem}
\usepackage{amsmath}
\usepackage{mathtools}
\usepackage{amsthm}
\usepackage{longtable}
\usepackage{comment}
\usepackage{indentfirst}

\usepackage{natbib}

\usepackage[utf8]{inputenc}
\usepackage[T1]{fontenc}
\usepackage{hyperref}
\usepackage{url}
\usepackage{booktabs}
\usepackage{amsfonts}
\usepackage{amssymb}
\usepackage{nicefrac}
\usepackage{microtype}
\usepackage{graphicx}
\usepackage{multirow}
\usepackage{array}
\usepackage{siunitx}
\usepackage{xcolor}
\usepackage{listings}
\usepackage{graphics}
\usepackage{subcaption}
\usepackage{caption}

\definecolor{codegreen}{rgb}{0,0.6,0}
\definecolor{codegray}{rgb}{0.5,0.5,0.5}
\definecolor{codepurple}{rgb}{0.58,0,0.82}
\definecolor{backcolour}{rgb}{0.97,0.97,0.97}

\title{\includegraphics[width=\textwidth]{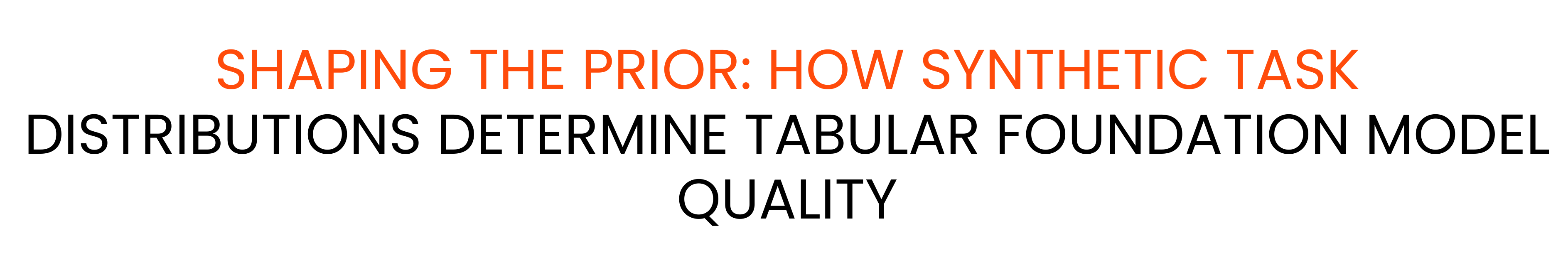}}

\author{
  Mohamed Bouadi$^*$, Nassim Bouarour$^*$,\\
  Varun Kulkarni, Shivam Dubey, Aditya Tanna, \\
  Vinay Kumar Sankarapu \\
  \affiliation{Lexsi Labs}\\
   \small{$^*$Equal contribution}
}

\setabstract{%
What determines the quality of a tabular foundation model? Unlike language or vision, tabular foundation models acquire their inductive biases almost entirely from synthetic pretraining distributions, yet the design of these distributions remains poorly understood. Standard synthetic priors are too well-behaved: they omit the irregularities and failure modes that determine deployment robustness. We introduce \textsc{O'Prior}, a compositional realism prior built around four coupled components: a hierarchical SCM meta-generator spanning diverse functional families; a modular realism engine covering heterogeneous marginals, missingness, and target transforms; an explicit stress module injecting confounding and support-query mismatch; and a curriculum-governed, leakage-safe generation protocol. To isolate prior design as the scientific variable, we hold architecture, optimizer, and compute budget fixed and vary only the synthetic task distribution. \textsc{O'Prior} yields consistent and substantial improvements in downstream accuracy and robustness across real tabular benchmarks, with gains concentrated in regimes characterized by distributional irregularities. Ablations confirm that mechanism diversity, realism composition, and shift-aware stress each contribute independently, their effects are not interchangeable. These results establish synthetic prior construction as a first-order and largely overlooked determinant of tabular foundation model quality.
}

\setkeywords{Tabular Foundation Models, Synthetic Prior Design, In-Context Learning, Structural Causal Models}

\runningtitle{Shaping the Prior: How Synthetic Task Distributions Determine Tabular Foundation Model Quality}

\begin{document}
\maketitle

\section{Introduction}
\label{sec:introduction}

Tabular data remains one of the most widely used modalities for predictive modeling, including healthcare, finance, industry, and sciences. Yet despite transformative advances in deep learning for language and vision~\cite{grinsztajn2022tree, mcelfresh2023neural}, gradient-boosted trees (GBT) remain the dominant approach on tabular benchmarks~\cite{chen2016xgboost, ke2017lightgbm, prokhorenkova2018catboost}, particularly under limited sample sizes, mixed feature types, and irregular statistical structure. This gap has motivated the emergence of new class of methods: tabular foundation models (TFMs). Recent systems such as \textsc{TabPFN}~\cite{tabpfn}, \textsc{TabPFN v2.5}~\cite{tabpfnv2.5}, \textsc{TabICL}~\cite{tabicl}, \textsc{OrionBix}~\cite{DBLP:conf/www/BouadiSTS26}, \textsc{OrionMSP}~\cite{orionmsp}, and \textsc{Limix}~\cite{limix} recast supervised tabular learning as amortized inference. After pretraining on a large collection of supervised tasks, the model receives a labeled context set at inference time and predicts query labels in a single forward pass without any dataset-specific training nor hyperparameter tuning. While this paradigm matches or exceeds strong GBTs baselines~\cite{tabarena} on standard benchmarks, these results raise a foundational question: what determines the quality of a pretrained tabular model?

Unlike language or vision foundation models, TFMs are not pretrained on naturally occurring data. Their inductive biases are induced almost entirely by synthetic tasks sampled from a hand-designed generative prior \cite{tabpfn}. The prior is therefore not merely a data source, it is the mechanism through which the model learns what tabular prediction problems look like. This makes prior design a first-order scientific question. Yet it has received almost no systematic study. Existing work demonstrates that synthetic pretraining can be effective, but does not establish \emph{which properties of the synthetic distribution are responsible} for downstream performance.

We argue this gap is consequential. Standard synthetic priors generate worlds that are too well-behaved: feature marginals are smooth, missingness is absent, support and query rows are exchangeable, and no spurious correlations exist. Real tabular datasets are shaped by confounding, selection effects, structured missingness, covariate shift, and shortcuts that break between training and test populations. A model pretrained on idealized synthetic tasks may learn useful generic predictors, but will not acquire the robustness that deployment requires.

This paper treats prior design as its primary scientific object. We introduce \textsc{O'Prior}, a compositional realism prior that generates tabular pretraining tasks through a hybrid SCM meta-generator, a modular realism engine, and an explicit distribution-shift and shortcut-stress module, all governed by a curriculum-controlled, leakage-safe generation protocol. To isolate the contribution of the prior, we hold architecture, optimizer, and compute budget fixed across all comparisons and vary only the synthetic task distribution. \textsc{O'Prior} yields consistent and substantial improvements over simpler baselines across real tabular benchmarks, with ablations showing that mechanism diversity, realism composition, and shift-aware stress each contribute independently.

\paragraph{Contributions.}
This paper makes the following contributions:
\begin{enumerate}[leftmargin=*, label=(\roman*)]
    \item We introduce \textsc{O'Prior}, a hierarchical compositional prior unifying structural causal diversity, modular realism perturbations, and controlled shift stress within a leakage-safe curriculum framework.\looseness=-1
    \item We establish a controlled evaluation protocol that isolates prior design as an independent variable, enabling clean attribution of downstream performance to specific generative choices.
    \item We provide systematic ablations identifying which realism components drive transfer quality, offering principled guidance for synthetic pretraining pipeline design.
\end{enumerate}

\section{\textsc{O’Prior}: Compositional realism priors for tabular pretraining}
\label{sec:O’Prior}
We introduce \textsc{O'Prior}, a compositional prior for TFMs pretraining implemented as a batched synthetic task generator. Its design rests on a simple but consequential observation: real tabular datasets are not clean samples from smooth input-output functions. They are shaped by latent causal structure, heterogeneous feature distributions, structured missingness, measurement artifacts, and distributional shifts between training and test populations. Standard synthetic priors address the first of these, functional diversity, but systematically neglect the rest. 

\begin{figure}[t]
    \centering
    \includegraphics[width=\linewidth, trim=45mm 0mm 45mm 0mm, clip]{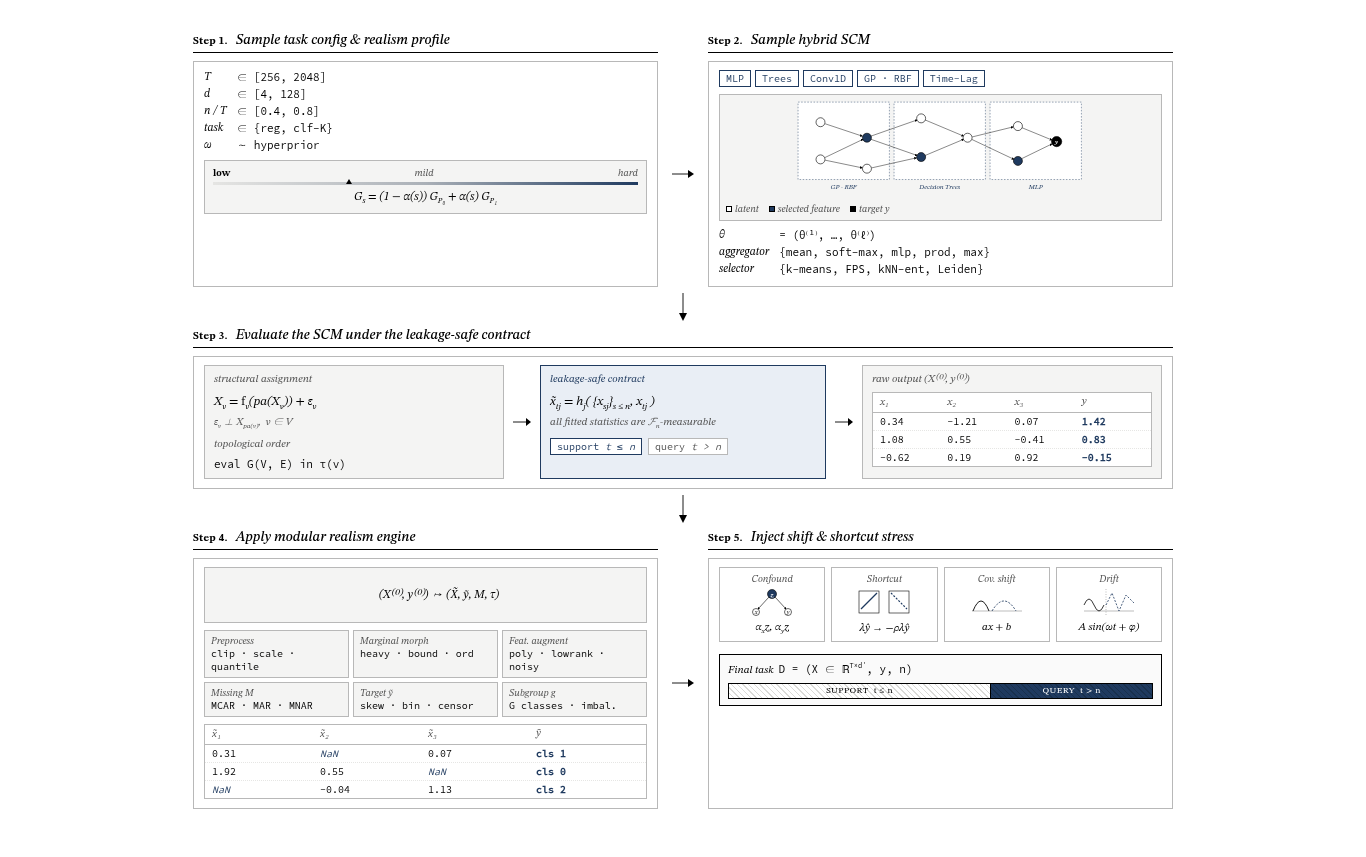}
    \caption{ \textbf{Schematic overview of the \textsc{O'Prior} compositional realism prior}. (i) Task-level hyperparameters and a curriculum-controlled realism profile $P \in \{\mathrm{LOW}, \mathrm{MILD}, \mathrm{HARD}\}$ are sampled jointly, with curriculum mixing  $G_s = (1-\alpha(s))G_{P_0} + \alpha(s)G_{P_1}$; (ii) A Hybrid SCM is sampled by composing base mechanisms from five families--MLPs, decision trees, one-dimensional convolutions, RBF Gaussian processes, and time-lagged VARs-- into a single random DAG, with per-node aggregation rules and a diversity-aware feature/target selector drawn jointly; (iii) The DAG is evaluated in topological order under the leakage-safe contract  $\tilde{x}_{tj} = h_j(\{x_{sj}\}_{s \leq n}, x_{tj})$, so every fitted statistic is $\mathcal{F}_n$-measurable; (iv) A modular realism engine maps $(X^{(0)}, y^{(0)})$ to $(\tilde{X}, \tilde{y}, M, \tau)$ through marginal morphing, feature augmentation, MCAR/MAR/MNAR missingness, target reshaping, and optional subgroup structure; (v) A shift-and-shortcut stress module injects latent confounding, in-support shortcuts that weaken or invert on query rows, affine covariate shift, and seasonal or regime drift, each controlled by independent Bernoulli gates and task-level parameters $\omega$.  The surviving episode is emitted as the final in-context task.}
    \textsc{10pt}
    \label{fig:oprior}
\end{figure}

\textsc{O'Prior} addresses this gap through a sequential generative pipeline as shown in Figure \ref{fig:oprior}. First, a structural causal model (SCM) generator samples a raw supervised task from various mechanisms. Second, a modular realism engine transforms this raw task to be more realistic by altering feature marginals, adding engineered structure, inducing missingness, and reshaping targets. Third, distribution-shifts, support-query mismatches, confounding, and spurious correlations are injected to harden in-context predictors against shortcuts. Finally, a curriculum-governed protocol controls the difficulty of these components over pretraining while enforcing leakage-safe transformations. Since these components are independent, \textsc{O'Prior} supports controlled attribution of downstream gains to specific generative choices.

\paragraph{Leakage-safe contract.} A TFM is evaluated as an in-context learner: it observes a labeled support set $\{(\mathbf{x}_t, y_t)\}_{t=1}^n$ and predicts labels for query rows $t > n$, which are unseen at inference time. Any preprocessing statistic computed over query rows would therefore constitute information leakage - it would give the generator knowledge the model cannot have at test time. Throughout \textsc{O'Prior}, we enforce the following contract: \emph{all fitted statistics are $\mathcal{F}_n$-measurable, computed exclusively from support rows}. Query rows are transformed only by functions of the support:
\begin{equation}
    \tilde{x}_{tj} = h_j\!\left(\{x_{sj}\}_{s \le n},\, x_{tj}\right), \quad t > n.
\end{equation}
We state this contract once here. All subsequent components respect it without re-statement.

\paragraph{Task format.} Each synthetic task is a supervised tuple $\mathcal{D} = (\mathbf{X} \in \mathbb{R}^{T \times d},\, \mathbf{y} \in \mathbb{R}^T,\, n)$, where rows $t \le n$ form the \emph{support} (labeled context) and rows $t > n$ form the \emph{query} set whose labels are predicted. The target $\mathbf{y}$ is either a continuous regression or a discrete classification target in $\{0, \ldots, K-1\}^T$. The generator produces complete in-context episodes rather than isolated examples.

\subsection{Structural causal model generator}
\label{sec:O’Prior-scm}
The first stage is responsible for \emph{functional and causal diversity}: it determines what variables exist, how they depend on one another, and how the target is generated from latent and observed causes. Without this stage, all downstream realism perturbations would be applied to a single impoverished family of mechanisms, limiting the prior's coverage of real-world predictive structures.

\paragraph{Prior and mechanisms.} Formally, a synthetic prior is a distribution $p(\theta)$ over a space of data-generating mechanisms $\Theta $. Each mechanism $\theta \in \Theta$ induces a distribution over supervised tasks: 
\begin{equation}
    \theta \sim p(\theta),
    \qquad
    \ (\mathbf{X}^{(0)},\mathbf{y}^{(0)}) \sim p_\theta(\mathbf{X},\mathbf{y}).
\end{equation}

In \textsc{O'Prior}, mechanisms are represented as structural causal models (SCMs) \cite{scm}: directed acyclic systems of structural assignments in which latent causes, observed covariates, and the response are generated from parent variables and independent exogenous noise.

\paragraph{SCM-based Models.} To cover the space of functional relationships found in real tabular data, \textsc{O'Prior} draws from five base mechanism classes. Some were introduced in prior work \cite{tabpfn,tabicl,limix}; others are introduced here.


\begin{itemize}
    \item {\it Multilayer perceptrons (MLPs)}: Previously proposed by~\cite{tabpfn}, it randomly initializes an $L$-layer MLP with Gaussian noise injected at each layer, and routes $k$ randomly initialized cause variables from which $\mathbf{X}$ and $\mathbf{y}$ are sampled from the hidden activations.

    \item {\it Decision Trees}: Structural equations are tree-based regressors fitted on-the-fly at each layer. We extend the original XGBoost-only formulation used by \cite{tabicl} to include Decision Tree~\cite{breiman2017classification}, Extra Trees~\cite{geurts2006extremely}, Random Forest~\cite{breiman2001random}, and a directly sampled variant.

    \item {\it Convolutional layers}: Structurally identical to MLPs but with $1$D convolutional layers of kernel size $k$ in place of fully connected layers, following \cite{limix}. This covers smooth local dependencies along a feature index.
    
    \item {\it Gaussian Process}: Each column of $\mathbf{X}$ is sampled from a GP with an RBF kernel via Cholesky factorization \cite{williams2006gaussian}. The target $\mathbf{y}$ is generated as a linear, quadratic, or MLP combination of the GP columns, covering smooth low-dimensional structure.\looseness=-1
    
    \item {\it Time Lagged}: A nonlinear vector autoregression of order $p$~\cite{terasvirta2010modelling} with randomly sampled weights and a linear projection as the target. This covers temporal and autoregressive causal structures. 

\end{itemize}

\paragraph{Hybrid SCM.}
Previous SCM models generate tasks through a single functional family. Real tabular datasets, however, are rarely so uniform: a financial dataset might exhibit linear relationships among macroeconomic indicators and polynomial interactions among other covariates, while a clinical dataset might combine tree-structured treatment rules with GP-like biological variation. To capture this, we introduce the \emph{Hybrid SCM}, a composition of multiple base classes within a single task.

A Hybrid SCM is defined as an ordered tuple $(\theta^{(1)}, \ldots, \theta^{(\ell)})$ of base mechanisms, where the output variables of $\theta^{(i)}$ may serve as inputs to $\theta^{(j)}$ for $j > i$. This defines a random directed acyclic graph whose edges correspond to different structural families. Node values are aggregated from their parents using one of five operators: mean, weighted softmax, MLP, product, or max, sampled per node. Features and targets are then selected from the full set of nodes using one of four diversity-aware strategies: \emph{k-means} clustering of node representations, \emph{farthest-point} greedy selection, \emph{entropy-based} ranking via k-nearest-neighbour entropy estimation~\cite{kraskov2004estimating}, and \emph{graph community detection} via the Leiden or Louvain algorithm~\cite{traag2019louvain}. These strategies ensure that the selected features span diverse regions of the representational space of the DAG rather than clustering around a single mechanism, which would undermine the heterogeneity the Hybrid SCM.

\subsection{Modular realism engine}
\label{sec:O’Prior-realism}

The SCM stage produces functionally diverse tasks, but its outputs remain idealized: feature marginals are approximately Gaussian, missingness is absent, and no observational artifacts are present. Real tabular datasets differ from this in ways that are not cosmetic - heavy tails, structured missing values, measurement noise, and heterogeneous feature types systematically affect which predictive signals a model can extract. The realism engine transforms each raw SCM output into an observational table that better reflects these properties. Conceptually, this separates the \emph{latent structural signal} produced by the SCM from the \emph{observational layer} introduced by data collection, measurement, and preprocessing.

We write the realism map as
\begin{equation}
    (\mathbf{X}^{(0)}, \mathbf{y}^{(0)})
    \longmapsto
    (\widetilde{\mathbf{X}}, \widetilde{\mathbf{y}}, \mathbf{M}, \boldsymbol{\tau}),
\end{equation}

where $\mathbf{M} \in \{0,1\}^{T \times d'}$ is a missingness mask and $\boldsymbol{\tau}$ stores per-column semantic type and imputation identity. All statistics are computed from support rows in accordance with the leakage-safe contract stated above.

\paragraph{Support-only preprocessing.} Real tabular pipelines almost never consume raw generative latents directly. Analysts and automated systems routinely rescale, compress extremes, and re-express skewed marginals before modeling; separately, real datasets exhibit measurement error, missing fields, and gross outliers \emph{on top of} whatever preprocessing has already occurred. We therefore use an initial \emph{support-consistent} preprocessing block for three reasons: (i) it stabilizes scales for later realism modules (e.g., heteroscedastic noise and missingness models defined on standardized signals); (ii) it matches the information constraints of ICL (the model must not rely on statistics that require peeking at the query split); and (iii) it yields a clearer scientific separation between \emph{structural} diversity (SCM) and \emph{observational} diversity (noise, missingness, tails).

Let
\[
    \mathbf{X}^{(0)}
    =
    \begin{bmatrix}
        \mathbf{X}^{\mathrm{sup}} \\
        \mathbf{X}^{\mathrm{q}}
    \end{bmatrix},
    \qquad
    \mathbf{X}^{\mathrm{sup}} \in \mathbb{R}^{n \times d}.
\]
For each column $j$, the generator computes support-only moments $\mu_j$ and $\sigma_j$, with a small floor on $\sigma_j$. These statistics define clipping thresholds
\[
    \tau_j^- = \mu_j - \kappa\sigma_j,
    \qquad
    \tau_j^+ = \mu_j + \kappa\sigma_j.
\]
We treat $\tau_j^\pm$ as an \emph{operational range} learned from the context: support rows may be winsorized or cleaned for extreme values, while query rows are clipped to the \emph{same} $[\tau_j^-,\tau_j^+]$ bounds. Next, optional monotone quantile maps and marginal transformations maybe be fitted on support rows and applied to both support and query rows. These include standardization, power transforms, skew correction, log-type shifts, and rank-based Gaussianization. This design enforces the same information constraint as in-context prediction: query rows may be transformed using functions fit on support rows, but not using query-label or global query-distribution information.

\paragraph{Feature augmentation and marginal morphing.} Real datasets often contain engineered features, redundant columns, and marginal distributions far from Gaussian. \textsc{O'Prior} addresses this in two steps. First, engineered columns may be appended: polynomial interactions, low-rank projections (with the projection basis estimated from the support and applied to queries), noisy duplicates, and random linear combinations of existing features. Second, selected columns undergo \emph{distributional morphing}: heavy-tailed variables (Student-t or Pareto-like tails), bounded variables (Beta or sigmoid maps), count-like variables (Poisson or negative-binomial), ordinalized variables (uniform or quantile bins), nonlinear CDF warps (e.g., Kumaraswamy), rank-Gaussian variables, heteroscedastic noise, redundant features, and sparse gross outliers. Together, these expose the pretrained model to the feature marginals it will encounter in real deployment, rather than the smooth distributions that SCMs naturally produce.

\begin{equation}
    \tilde{x}_{tj} = x_{tj} + \sigma_j \exp(\alpha |z_{tj}|)\,\epsilon_{tj}, \quad \epsilon_{tj} \sim \mathcal{N}(0,1),
\end{equation}

\paragraph{Structured missingness.} Missing values are ubiquitous in real tabular data and arise through distinct mechanisms that interact differently with predictive models. For each column $j$, \textsc{O'Prior} samples a mechanism $R_j \in \{\mathrm{MCAR}, \mathrm{MAR}, \mathrm{MNAR}\}$ with mixture weights from $\boldsymbol{\omega}$. Under MCAR, entries are missing independently with rate $\pi_j$. Under MAR, missingness depends on a driver column $k(j) \neq j$:\looseness=-1
\begin{equation}
    \mathbb{P}(m_{tj} = 1 \mid \mathbf{x}_t) = \sigma\!\left(\beta_{\mathrm{MAR}}\, z_{t,k(j)} + b_j\right),
\end{equation}
with $b_j$ calibrated from a target support rate. Under MNAR, missingness depends on the feature's own value:
\begin{equation}
    \mathbb{P}(m_{tj} = 1 \mid x_{tj}) = \sigma\!\left(\beta_{\mathrm{MNAR}}\, s_j\, z_{tj} + b_j\right),
\end{equation}
where $s_j \in \{-1, +1\}$ is derived from support skewness. Missing entries are imputed using support-only statistics - mean, median, constant fills, sampled support values, or Gaussian noise around the support mean - with the imputation strategy stored as column metadata so the model can condition on it when metadata is provided.

\paragraph{Target transformations.} The raw target $\mathbf{y}^{(0)}$ is transformed to diversify the shape of the prediction problem. For regression, support-only normalization is applied first:
\begin{equation}
    \tilde{y}_t = \frac{y_t - \bar{y}}{s_y}, \quad \bar{y},\, s_y \text{ from } t \le n,
\end{equation}
followed by optional skew transforms, heavy-tailed noise, mixture perturbations, bounded reparameterizations, or censoring-style transformations that model survival outcomes. For classification, a real-valued latent score is discretized into $K$ classes using support quantiles or ranks; class labels may be permuted and label noise may be introduced by randomly flipping a fraction of labels, producing tasks with variable class balance, margin structure, and label reliability.

\paragraph{Subgroup structure.} Real datasets frequently contain subgroups - demographic, institutional, or experimental - that induce systematic variation in the target beyond the primary predictive signal. \textsc{O'Prior} optionally adds a categorical subgroup attribute $g_t \in \{1, \ldots, G\}$ with balanced or imbalanced group probabilities, modifying the target through group-specific intercepts, random projections of one-hot encodings, or interactions with continuous features. Domain-specific extensions, such as finance-style volatility shocks, are available as independently ablatable modules.

\subsection{Distribution Shift and Shortcut Stress}
\label{sec:O’Prior-shift}
The realism engine improves the marginal and observational properties of each task, but it does not alter the relationship between support and query rows. In practice, however, support and query populations routinely differ - through confounding, covariate shift, temporal drift, or spurious predictors that are informative in training but unreliable at test time. For an in-context learner, these are not edge cases; they are the primary failure modes. This stage injects structured support-query mismatch to expose the model to exactly these conditions during pretraining. All shift strengths are drawn from $\boldsymbol{\omega}$ and controlled by task-level Bernoulli switches, enabling clean isolation of each mechanism in ablations.

\paragraph{Latent confounding.} To model shared unobserved causes, the generator samples a latent vector $\mathbf{z} \in \mathbb{R}^T$, normalized to zero mean and unit variance on support rows. For a random subset of columns $\mathcal{J}_{\mathrm{cf}}$, it applies:
\begin{equation}
    \tilde{x}_{tj} \leftarrow \tilde{x}_{tj} + \alpha_x z_t, \quad j \in \mathcal{J}_{\mathrm{cf}}, \qquad \tilde{y}_t \leftarrow \tilde{y}_t + \alpha_y z_t,
\end{equation}

with $\alpha_x, \alpha_y$ proportional to a global confounding strength in $\boldsymbol{\omega}$. The shared factor $z_t$ induces feature-target dependence that is not a direct causal effect, requiring the model to distinguish genuine predictive signal from confounded association.

\paragraph{Spurious support predictors.} To stress-test shortcut reliance, selected columns are made highly predictive on support rows but unreliable on query rows. Using support-normalized labels $\check{y}_t = (y_t - \bar{y})/s_y$, support entries for $j \in \mathcal{J}_{\mathrm{sp}}$ are set as: 
\begin{equation}
    \tilde{x}_{tj} = \lambda\,\check{y}_t + \epsilon_{tj}, \quad t \le n,
\end{equation}

while query entries obey: 
\begin{equation}
    \tilde{x}_{tj} = s_j\,\rho\,\lambda\,\check{y}_t + \epsilon_{tj}, \quad t > n,
\end{equation} 

where $s_j \in \{-1,+1\}$ may flip the association and $\rho \in (0,1]$ weakens it. These columns are strong shortcuts in-support but degraded or inverted out-of-support, a direct stress test for models that over-rely on spurious in-context correlations.

\paragraph{Covariate shift.} For a subset $\mathcal{J}_{\mathrm{cs}}$, query rows undergo affine perturbation:
\begin{equation}
    \tilde{x}_{tj} \leftarrow a_j\,\tilde{x}_{tj} + b_j, \quad t > n,\; j \in \mathcal{J}_{\mathrm{cs}},
\end{equation}

with $(a_j, b_j)$ scaled relative to support dispersion. This shifts the marginal distribution of query features away from the support without altering the causal task structure.

\paragraph{Temporal and seasonal drift.} For sequence-structured tasks, the generator introduces non-stationarity via two mechanisms. A seasonal component adds a periodic signal to the target:
\begin{equation}
\Delta y_t = A\sin(\omega t + \phi), \quad A \propto s_y.
\end{equation}

Regime drift is produced by sampling one or more change points and applying segment-specific affine transforms to selected columns, with sigmoid or abrupt blending at boundaries. Together, these cover the temporal distribution shift patterns common in financial, clinical, and sensor tabular data.

\subsection{Curriculum and Leakage-Safe Generation}
\label{sec:O’Prior-protocol}

The three stages above define a rich family of synthetic tasks, but applying maximal realism uniformly from the start of pretraining is not necessarily optimal. Early in training, the model has not yet learned basic predictive structure; exposing it immediately to heavy confounding, severe missingness, and adversarial shortcuts may slow convergence or prevent the acquisition of fundamental in-context learning behavior. This motivates a curriculum: staging realism difficulty over pretraining so that the model builds competence progressively. Critically, the curriculum operates entirely on the prior, the model architecture, optimizer, and compute budget remain fixed throughout, making curriculum schedule an independently ablatable experimental axis.

\paragraph{Realism profiles.} \textsc{O'Prior} defines three realism presets $\mathcal{P} \in \{\textsc{low}, \textsc{mild}, \textsc{hard}\}$. Each induces a distinct hyperprior $G_{\mathcal{P}}$ over the same parameterization by restricting the support of key coordinates of $\boldsymbol{\omega}$ - missingness rates, tail severity, confounding strength, spurious feature fraction, and shift magnitude become progressively less constrained from \textsc{low} to \textsc{hard}. Because the parameterization is shared, profiles can be directly compared or continuously interpolated.

\paragraph{Curriculum over profiles.} Let $s$ denote the synthetic batch index and $\alpha(s) \in [0,1]$ a curriculum schedule (linear, cosine, or step) over a warmup horizon $S_0$. The task hyperprior at step $s$ is: 
\begin{equation}
    G_s = (1 - \alpha(s))\,G_{\mathcal{P}_0} + \alpha(s)\,G_{\mathcal{P}_1}.
\end{equation}

The interpolation mixes discrete choice supports and linearly interpolates numeric ranges for rates, strengths, and probabilities. Early batches therefore emphasize simpler mechanisms and mild observational artifacts; later batches introduce harder missingness, heavier tails, stronger shifts, and more adversarial shortcuts.

\paragraph{Quality control.} Generated tasks are rejected and resampled when they fail basic validity checks: too few active features after pruning near-constant columns, collapsed or severely imbalanced support classes in classification tasks, or degenerate target variation in regression tasks. These checks are deliberately minimal, they prevent pathological draws without over-filtering the prior toward an unrealistically clean task distribution.

\section{Experimental Evaluation}
We provide an empirical analysis about the contribution of each component of \textsc{O’Prior}. We create different variants of the prior which are defined by combinations of its components and compare them to state-of-the-art priors, for tabular classification, and relying on a real-world playground. 

\subsection{Experimental Protocol}
\label{sec:protocol}

\paragraph{Controlled evaluation principle.} The central methodological commitment of this work is isolation: we hold the model architecture, optimizer, training budget, and evaluation pipeline fixed across all conditions and vary only the synthetic task distribution used for pretraining. Any observed difference in downstream performance is therefore attributable to the prior alone, not to differences in capacity, compute, or hyperparameter tuning.

\paragraph{Model and training infrastructure.} \textsc{nanoTabPFN} \cite{nanotabpfn} as our base TFM -- a lightweight, fully open-source reimplementation of the \textsc{TabPFN v2} \cite{tabpfnv2} architecture designed as a research testbed for studying TFMs under controlled conditions. All experiments use the TFM-Playground training protocol\footnote{https://github.com/automl/TFM-Playground/}, which provides a unified interface for loading synthetic prior data and evaluating the resulting model. \textsc{nanoTabPFN} is used deliberately: at reduced scale, pretraining requires only tens of thousands of synthetic datasets rather than millions, making controlled prior ablations computationally feasible without sacrificing the qualitative conclusions.

All models are trained under an identical budget: $40,000$ synthetic datasets per prior, each of size $T \in [512, 1024]$ rows and $d \in [3, 50]$ features, with a batch size of $4$ tables per optimization step, $1,000$ steps per epoch, and $10$ epochs of training, without hyperparameter tuning between conditions.

\paragraph{Baselines.} We compare against three existing generators under the same training budget and model:
\begin{itemize}
    \item \textbf{TabPFN v1} generator: the original SCM-based prior from \cite{tabpfn}, which generates synthetic datasets using MLP-based SCMs with randomly initialized Bayesian neural network weights and Gaussian noise injected at each layer. It provides functional diversity but no realism perturbations, no structured missingness, and no distribution shift.

    \item \textbf{TabICL-v1} generator \cite{tabicl}: extends the MLP-based SCM prior with tree-based structural equations (XGBoost), producing tasks whose decision boundaries better reflect the inductive biases of gradient-boosted trees.

    \item \textbf{TabICL-v2} generator \cite{tabiclv2}: augments TabICL-v1 with additional SCM diversity and a richer set of feature-level transforms, including improved handling of categorical features and target distributions. Relative to v1, it broadens functional coverage while remaining free of explicit realism perturbations and shift mechanisms.

\end{itemize}

\paragraph{O'Prior ablation groups.} To attribute performance gains to specific generative components, we construct nine variants organized into four groups following the sequential pipeline of Section~\ref{sec:O’Prior}. We use the following shorthand: \textit{SM} (basic SCM families), \textit{SH} (Hybrid SCM), \textit{MR} (moderate realism), \textit{SR} (strong realism), \textit{SD} (distribution shift and shortcut stress). Groups 1--3 each isolate a single stage by enabling it while holding the others off; Group 4 combines all components with a linear mild-to-hard realism curriculum and serves as the full \textsc{O'Prior} configuration. This design supports three types of comparison: baselines vs \textsc{O'Prior} variants, within-group ablation to isolate individual components, and cross-group comparison to assess interactions. Table~\ref{tab:ablation_variants} summarizes all variants.

\begin{table}[ht]
\centering
\caption{\textsc{O'Prior} ablation variants. G4 is the complete \textsc{O'Prior} configuration; all other variants isolate specific generative components to enable controlled attribution.}
\textsc{10pt}
\label{tab:ablation_variants}
\begin{tabular}{lll}
\hline
Variant & Components & Purpose \\ \hline
G1a & SM & SCM diversity without Hybrid \\
G1b & SH & Hybrid SCM only \\
G1c & SM + SH & Full SCM diversity \\ \hline
G2a & SM + MR & Effect of moderate realism \\
G2b & SM + SR & Effect of strong realism \\
G2c & SM + SH + SR & Realism with full SCM diversity \\ \hline
G3a & SM + SD & Shift stress without realism \\
G3b & SM + SH + SD & Shift stress with full SCM diversity \\ \hline
G4 & SM + SH + SR + SD + Curriculum & Full \textsc{O'Prior} pipeline \\ \hline
\end{tabular}
\end{table}
    
\paragraph{Benchmarks.} We evaluate on two complementary benchmarks covering $52$ tabular classification tasks in total: (i) \textit{TabArena v0.1} \cite{tabarena} -- which is included in the TFM-Playground, comprising $21$ curated tabular classification datasets with documented preprocessing and evaluation splits; (ii) \textit{OpenML-CC18 (filtered)} -- we sample all classification datasets from the OpenML-CC18 benchmark \cite{openmlcc18} satisfying: $d \in [2, 50]$ features and $N < 10{,}000$ samples, yielding $31$ datasets. This subset targets the small-to-medium data regime where in-context tabular learners are most competitive. Full dataset IDs and task specifications are provided in Appendix~\ref{app:datasets}.

\paragraph{Metrics.} We report ROC-AUC, accuracy (ACC), and macro-averaged F1 score on held-out test splits for all datasets and conditions. For multi-class tasks, ROC-AUC is computed using the one-vs-rest formulation. Results are averaged across datasets within each benchmark.

\subsection{Main Results}
\label{sec:main_results}
\subsubsection{Ablation Analysis}
\begin{table}[!t]
\small
  \centering
  \caption{Avg ROC-AUC, Accuracy, and F1-score for different compositions and variants of \textsc{O’Prior}.}
  \textsc{10pt}
  \label{tab:results}
  \resizebox{\linewidth}{!}{%
  \begin{tabular}{l lcccccc}
    \toprule
    \multirow{2}{*}{Group} & \multirow{2}{*}{Variants} 
    & \multicolumn{3}{c}{TabArena-v0.1} 
    & \multicolumn{3}{c}{OpenML-CC18} \\ 
    \cmidrule(lr){3-5} \cmidrule(lr){6-8}
    & & ROC-AUC & ACC & F1-score & ROC-AUC & ACC & F1-score \\ 
    \midrule
    \multirow{3}{*}{Baselines}
    & TabPFN    & 0.5758 & 0.7257 & 0.3724 & 0.5189 & 0.5770 & 0.3105 \\ 
    & TabICL-v1 & 0.6361 & 0.7366 & 0.3842 & 0.5754 & 0.5883 & 0.3139 \\ 
    & TabICL-v2 & 0.7910 & 0.8026 & 0.5234 & 0.7240 & 0.6402 & 0.4442 \\ 
    \midrule
    \multirow{3}{*}{SCM/Hybrid}
    & SM (G1a)    & 0.7881 & 0.8087 & 0.5555 & 0.7411 & 0.6641 & 0.4929 \\
    & SH (G1b)    & \textbf{\underline{0.8335}} & 0.8349 & \textbf{\underline{0.6148}} & 0.8228 & 0.7299 & 0.5789 \\ 
    & SM+SH (G1c) & 0.8324 & 0.8298 & 0.5879 & \textbf{\underline{0.8313}} & \textbf{\underline{0.7427}} & \textbf{\underline{0.5906}} \\ 
    \midrule
    \multirow{3}{*}{Realism}
    & SM+MR (G2a) & 0.8102 & 0.8152 & 0.5699 & 0.7967 & 0.7089 & 0.5473 \\
    & SM+SR (G2b) & 0.8315 & 0.8331 & \underline{0.6143} & 0.8128 & 0.7284 & 0.5848 \\ 
    & SM+SH+SR (G2c) & 0.8295 & \textbf{\underline{0.8366}} & 0.6138 & 0.8011 & 0.7045 & 0.5395 \\
    \midrule
    \multirow{2}{*}{Distribution Shifts}
    & SM+SD (G3a) & 0.8012 & 0.8128 & 0.5400 & 0.7806 & 0.6969 & 0.5023 \\
    & SM+SH+SD (G3b) & 0.8192 & 0.8250 & 0.5725 & 0.8064 & 0.7170 & 0.5271 \\
    \midrule
    \multirow{1}{*}{Curriculum}
    & G4 
    & 0.8194
    & 0.8212
    & 0.5829 
    & 0.8245
    & 0.7239
    & 0.5693\\
    \bottomrule
  \end{tabular}
  }
\end{table}

\paragraph{Overview and Motivation.} The ablation study is designed to answer a specific and narrow question: given a fixed architecture and a fixed training budget, how much does the choice of synthetic prior matter, and which components of that prior are responsible for downstream gains? This framing is important because prior work conflates architectural improvements with prior improvements, \textsc{TabPFN V2}, for instance, enriched both simultaneously, making attribution impossible. Here, the prior is the only variable.

The nine \textsc{O'Prior} variants are organized into four sequential groups that mirror the pipeline introduced in Section~\ref{sec:O’Prior}: Group 1 targets structural mechanism diversity (SCM and Hybrid SCM), Group 2 adds the realism engine on top of a fixed SCM base, Group 3 adds the distribution-shift and shortcut-stress module, and Group 4 combines all components under a curriculum schedule. Three baselines anchor the comparison at the low end.

Table~\ref{tab:results} reports average ROC-AUC, accuracy, and macro F1-score for all conditions across both benchmarks.

\paragraph{Baseline Progression.} The three baselines already reveal that prior design matters substantially, even before any \textsc{O'Prior} variant is considered. \textsc{TabPFN}, the oldest and simplest prior, obtains the weakest performance across both benchmarks, with ROC-AUC of 0.576 on TabArena and 0.519 on OpenML-CC18, and macro F1 below 0.37 on both. \textsc{TabICL-v1}, which extends the MLP-based SCM with XGBoost-based structural equations, improves modestly over \textsc{TabPFN} across all metrics. The improvement is consistent but small, suggesting that adding tree-based mechanisms to an otherwise unchanged prior delivers marginal gains in this budget-constrained setting. \textsc{TabICL-v2} represents a more substantial advance, reaching ROC-AUC of 0.791 on TabArena and 0.724 on OpenML-CC18, with macro F1 improving by approximately 14 points over TabICL-v1 on TabArena. This progression across the three baselines confirms the central hypothesis: richer synthetic task distributions yield better in-context learners, and the prior is a first-order determinant of model quality.

\paragraph{Structural Mechanism Diversity.} Group 1 tests the contribution of the SCM generator stage before any realism or shift perturbations are applied. The basic SCM prior (G1a) already matches \textsc{TabICL-v2} on TabArena ROC-AUC (0.788 vs 0.791) and exceeds it on F1 (0.556 vs 0.523), while outperforming it on OpenML-CC18 across all three metrics. This is a notable result: diversifying the space of structural mechanisms, without any other modification, is sufficient to match or surpass a substantially more complex baseline.

The Hybrid SCM alone (G1b) delivers the largest single-component gain in the ablation. Adding compositional multi-mechanism tasks -- where different structural families generate subsets of variables within the same DAG -- pushes TabArena ROC-AUC to 0.834 and macro F1 to 0.615, the highest F1 of any individual variant on TabArena. On OpenML-CC18, G1b reaches 0.823 ROC-AUC and 0.579 F1, representing a 10-point F1 improvement over \textsc{TabICL-v2}. Combining basic and Hybrid SCMs (G1c) gives the strongest OpenML-CC18 ROC-AUC among all non-curriculum variants (0.831) and the highest OpenML-CC18 accuracy (0.743), suggesting that the two mechanism families are complementary: the basic SCM families cover distinct functional regions that the Hybrid SCM alone may undersample.

The Group 1 finding is perhaps the most consequential result in the ablation: mechanism diversity alone, with no realism perturbations and no shift stress, accounts for the majority of the gap between the weakest baseline (\textsc{TabPFN}) and the best \textsc{O'Prior} configurations. The SCM generator is the primary source of transfer improvement.

\paragraph{Modular Realism Engine.} Group 2 adds the realism engine on top of the basic SCM prior (SM), holding the structural mechanisms fixed. The moderate-realism variant (G2a, SM+MR) improves over SM on all metrics on both benchmarks. This confirms that realism perturbations, heterogeneous marginals, structured missingness, noisy targets, and target-shape variation, are beneficial rather than merely adding noise. The model learns something transferable from harder observational artifacts.

Strong realism (G2b, SM+SR) consistently outperforms moderate realism across all six metric-benchmark combinations. TabArena F1 rises from 0.570 to 0.614, and OpenML-CC18 F1 from 0.547 to 0.585. The monotone relationship between realism severity and performance suggests that the model benefits from increased exposure to the kinds of distributional irregularities it encounters in real datasets, heavy-tailed marginals, MAR and MNAR missingness patterns, label noise, and bounded or skewed target distributions.

The interaction between Hybrid SCM and strong realism (G2c, SM+SH+SR) reveals a nontrivial dependency on training budget. While G2c improves over the basic SCM prior on TabArena, it does not uniformly dominate G2b on OpenML-CC18, where its ROC-AUC (0.801) is lower than G2b (0.813) and its accuracy (0.705) falls below G2b (0.728). The interpretation is that combining maximal structural diversity with maximal realism perturbations places a higher learning burden on the model: under a fixed budget of 40,000 synthetic datasets, the model may not fully amortize the increased variability of both stages simultaneously. This interaction is not a failure of either component; it is a budget-dependent constraint. With larger pretraining budgets, G2c would be expected to dominate.

\paragraph{Distribution Shift and Shortcut Stress.} Group 3 isolates the contribution of the shift stress module: latent confounding, spurious support predictors, covariate shift, and temporal drift. Adding the shift module to the basic SCM prior (G3a, SM+SD) improves substantially over SM across both benchmarks. Adding the Hybrid SCM alongside the shift module (G3b, SM+SH+SD) further improves all metrics, reaching TabArena ROC-AUC of 0.819 and OpenML-CC18 F1 of 0.527.

The average- case gains from the shift module are smaller than those from the realism engine, which is expected by design. The shift module is not primarily intended to improve average i.i.d. accuracy; it is designed to expose the model to support-query mismatch and shortcut failure modes that manifest most clearly under distributional irregularity. Its contribution is therefore best understood as a robustness-oriented component. The benefit will be most pronounced when the downstream evaluation explicitly includes shifted or out-of-distribution test conditions -- a regime that the standard benchmarks used here do not fully capture. Under more targeted robustness evaluations, the shift module's contribution would likely rank higher.

\paragraph{Full Curriculum.} The curriculum variant (G4) combines all \textsc{O'Prior} components -- hybrid SCMs, strong realism, shift stress -- under a linear mild-to-hard schedule that begins with the LOW realism profile and transitions to HARD over the warmup horizon. It substantially outperforms all three baselines: over \textsc{TabICL-v2}, G4 improves TabArena ROC-AUC by 2.8 points, accuracy by 1.9 points, and F1 by 5.9 points. On OpenML-CC18, the improvements are 10.1, 8.4, and 12.5 points respectively.

However, G4 does not dominate the best static configurations within the \textsc{O'Prior} ablation. Under the fixed budget of 40,000 synthetic datasets and 10 epochs, configurations such as G1b, G1c, and G2b outperform G4 on several metric-benchmark combinations. The explanation is structural: under a fixed step budget, spending early updates on simpler LOW-profile tasks reduces the number of updates drawn from the target HARD distribution. Static strong priors allocate every gradient step to the final task distribution, making them more sample-efficient when the budget is the binding constraint. Curriculum is not a universally superior strategy; it is a schedule-dependent design choice. Its advantage materializes when optimization stability -- not sample efficiency -- is the bottleneck, or when the total pretraining budget is large enough that the warmup phase does not materially reduce exposure to the hardest regime.

\paragraph{Summary of Ablation Findings.} Three conclusions emerge consistently across all conditions.

\begin{itemize}
    \item First, structural mechanism diversity is the dominant source of transfer improvement. The Group 1 variants outperform all baselines and account for the majority of the performance gap over \textsc{TabICL-v2}, despite using no realism perturbations or shift mechanisms. The Hybrid SCM, which composes multiple functional families within a single synthetic task, is the single most impactful component in the ablation.

    \item Second, realism perturbations and shift stress add complementary, independent gains. Adding the realism engine to a fixed SCM prior consistently improves all metrics, with stronger realism monotonically outperforming moderate realism. The shift stress module similarly improves over the SCM-only prior. Critically, the ablation design -- with each component toggled independently -- confirms that these effects are not interchangeable: mechanism diversity and realism each capture qualitatively distinct aspects of what makes a synthetic prior useful.

    \item Third, curriculum is a schedule-dependent design choice rather than a uniformly superior strategy. Under short fixed budgets, static strong priors outperform the curriculum configuration because every training step is drawn from the target distribution. Curriculum becomes the right choice when optimization stability is the limiting factor or when training budgets are sufficiently large that the warmup cost is amortized.
\end{itemize}

\subsubsection{Prior Quality Assessment via Structural Alignment}

Downstream benchmark performance measures the effect of a synthetic prior after it has been absorbed by a trained tabular foundation model. This is the primary evaluation criterion in our study, but it does not directly answer a complementary question: do the synthetic tables generated by the prior exhibit structural properties that resemble real tabular data? We therefore evaluate the synthetic generators themselves using a schema-agnostic adaptation of TabStruct~\citep{jiang2025tabstruct}. The goal is not to reproduce any single real dataset, but to assess whether a prior generates tables with realistic marginal behavior and inter-feature dependency structure.

\paragraph{Schema-agnostic adaptation.}
TabStruct is originally designed for fixed-schema synthetic data evaluation, where synthetic and real tables share aligned columns. This assumption is incompatible with \textsc{O'Prior}: by construction, \textsc{O'Prior} generates variable-schema supervised tasks with varying numbers of rows, features, and causal mechanisms. Column-wise matching would therefore penalize precisely the form of diversity that the prior is designed to produce. We instead compare generators through pooled, schema-agnostic statistics. For each real reference dataset, we sample synthetic tables from Full \textsc{O'Prior} and \textsc{TabICL-v2}, then evaluate two complementary properties: one-dimensional marginal alignment and correlation-spectrum alignment. This adaptation preserves the spirit of TabStruct while making the evaluation appropriate for priors that define distributions over tasks rather than replicas of a fixed table.

\paragraph{Evaluation protocol.}
We evaluate against four real tabular reference datasets spanning different dimensionalities and domains: Electricity (OpenML ID:151), Diabetes (OpenML ID:37), Telco Churn (OpenML ID:40701), and Home Credit. Electricity and Diabetes provide lower-dimensional references, while Churn and Home Credit contain richer inter-feature structure and therefore provide a stronger test of structural realism. For each reference dataset and each generator, we run 10 independent Monte Carlo iterations. In each iteration, we sample 50 synthetic tables from the generator and compute the structural-alignment metrics described below. Results are reported as mean $\pm$ standard deviation across iterations.

\paragraph{Metrics.}
We compute two scores per iteration. The marginal score measures alignment between the pooled feature-value distribution of the synthetic tables and that of the real reference. Feature values from synthetic tables are pooled, capped by reservoir sampling when necessary, quantile-normalized to $[0,1]$, and compared to the reference using the first Wasserstein distance:
\begin{equation}
    S_{\mathrm{marginal}} = \exp(-W_1).
\end{equation}

The correlation score measures structural alignment. For each synthetic table, we compute the eigenvalue spectrum of its feature-correlation matrix, pool spectra across sampled tables, and compare the resulting distribution to the reference spectrum:
\begin{equation}
    S_{\mathrm{corr}} = \exp(-5W_1).
\end{equation}

The scaling factor makes the correlation score comparable in range to the marginal score. Finally, we combine the two dimensions using the geometric mean:
\begin{equation}
    Q = \sqrt{S_{\mathrm{marginal}} S_{\mathrm{corr}}}.
\end{equation}

This composite score penalizes generators that perform well on only one dimension. A high value of $Q$ therefore requires both realistic marginal behavior and realistic dependency structure.

\paragraph{Results.}

\begin{table}[t]
\centering
\caption{TabStruct-adapted realism evaluation results. Scores are reported as mean $\pm$ standard deviation over 10 independent Monte Carlo iterations.}
\textsc{10pt}
\label{tab:tabstruct}
\begin{tabular}{llccc}
\toprule
Reference Data & Synthetic & Marginal & Correlation & $Q$ \\
\midrule
Electricity (ID:151) & Full {\sc O'Prior} & $0.979 \pm 0.003$ & $0.795 \pm 0.027$ & $0.882 \pm 0.015$ \\
Electricity (ID:151) & \textsc{TabICL}-v2 & $0.995 \pm 0.000$ & $0.760 \pm 0.025$ & $0.870 \pm 0.014$ \\
\midrule
Diabetes (ID:37) & Full {\sc O'Prior} & $0.974 \pm 0.003$ & $0.745 \pm 0.025$ & $0.852 \pm 0.015$ \\
Diabetes (ID:37) & \textsc{TabICL}-v2 & $0.988 \pm 0.002$ & $0.710 \pm 0.026$ & $0.838 \pm 0.015$ \\
\midrule
Churn (ID:40701) & Full {\sc O'Prior} & $0.971 \pm 0.003$ & $0.910 \pm 0.013$ & $0.940 \pm 0.007$ \\
Churn (ID:40701) & \textsc{TabICL}-v2 & $0.985 \pm 0.000$ & $0.887 \pm 0.021$ & $0.934 \pm 0.011$ \\
\midrule
Home Credit & Full {\sc O'Prior} & $0.957 \pm 0.002$ & $0.941 \pm 0.006$ & $0.949 \pm 0.004$ \\
Home Credit & \textsc{TabICL}-v2 & $0.968 \pm 0.001$ & $0.906 \pm 0.028$ & $0.937 \pm 0.015$ \\
\bottomrule
\end{tabular}
\end{table}

Table~\ref{tab:tabstruct} reports the adapted TabStruct scores. \textsc{O'Prior} achieves the highest composite quality score on all four reference datasets. The gains are most pronounced on the structurally richer references: on Churn, \textsc{O'Prior} obtains $Q=0.940 \pm 0.007$ compared with $0.934 \pm 0.011$ for \textsc{TabICL-v2}, while on Home Credit it obtains $0.949 \pm 0.004$ compared with $0.937 \pm 0.015$. \textsc{O'Prior} also has substantially lower variance on Home Credit, indicating that its structural realism is more stable across Monte Carlo samples.

The source of this improvement is correlation alignment. \textsc{O'Prior} obtains higher correlation scores on every reference dataset: $0.795$ vs.\ $0.760$ on Electricity, $0.745$ vs.\ $0.710$ on Diabetes, $0.910$ vs.\ $0.887$ on Churn, and $0.941$ vs.\ $0.906$ on Home Credit. This pattern directly supports the design motivation of \textsc{O'Prior}. Its compositional SCM generator does not merely diversify univariate feature shapes; it produces structured dependencies between variables through heterogeneous causal mechanisms, shared latent structure, subgroup effects, and support-query perturbations. These mechanisms are exactly what should improve correlation-spectrum realism.

\textsc{TabICL-v2} obtains higher marginal scores across all four references. We do not interpret this as evidence of stronger overall realism. \textsc{TabICL-v2} produces fixed-format synthetic tables, so pooling many homogeneous samples yields very stable one-dimensional marginal estimates. \textsc{O'Prior} intentionally generates variable-schema tasks with heterogeneous feature counts, mechanisms, and marginal transformations. This broader coverage can increase the Wasserstein distance under a pooled marginal metric, even when the generated tasks are more useful and structurally realistic. The composite score and correlation score are therefore more informative for evaluating variable-schema priors, because they capture whether the generator produces realistic tabular structure rather than whether it matches a single fixed schema.

\paragraph{Structural visualization.}
\begin{figure}[t]
    \centering
    \includegraphics[width=\linewidth]{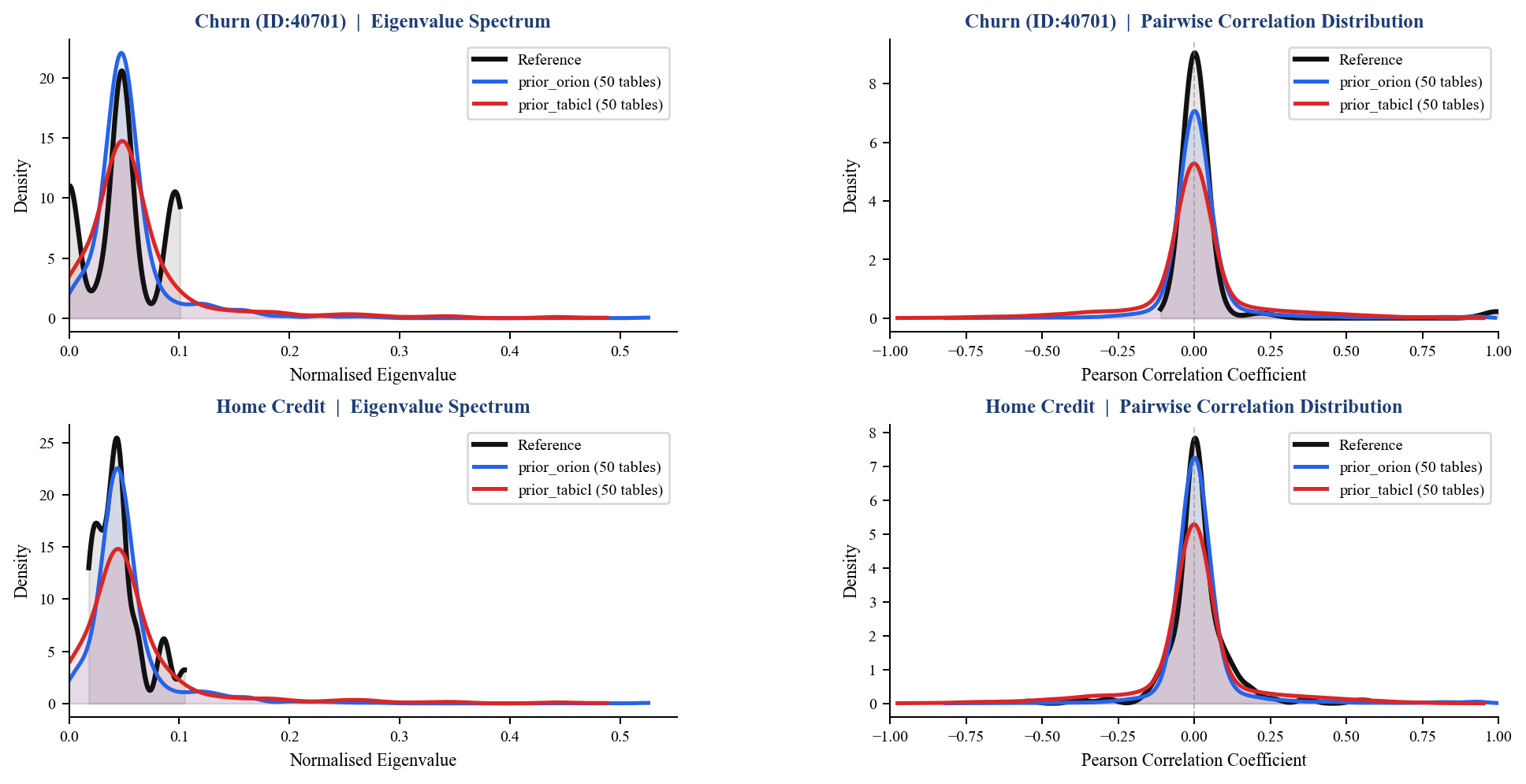}
    \caption{
    Structural alignment between real reference datasets and synthetic priors. Eigenvalue spectra and pairwise-correlation distributions are aggregated over 50 sampled tables per synthetic source. 
    \textsc{O'Prior} better matches the eigenvalue decay and correlation spread of high-dimensional real datasets, while \textsc{TabICL}-v2 produces weaker inter-feature dependence.}
    \textsc{10pt}
    \label{fig:eigenspectra}
\end{figure}

The visualization in Figure~\ref{fig:eigenspectra} provides qualitative support for the numerical results. The eigenvalue-spectrum plots show that \textsc{O'Prior} more closely follows the decay pattern of real reference datasets, especially on Churn and Home Credit, where real tables exhibit stronger inter-feature dependence. In contrast, \textsc{TabICL-v2} produces spectra closer to weakly dependent or overly uniform feature sets. The pairwise-correlation visualizations show the same effect from a complementary perspective: \textsc{O'Prior} generates broader and more realistic correlation structure, while \textsc{TabICL-v2} is more concentrated near zero correlation. These visual patterns explain why \textsc{O'Prior} consistently improves $S_{\mathrm{corr}}$ and, consequently, the composite score $Q$.

Overall, this experiment shows that \textsc{O'Prior} improves the quality of the synthetic prior at the data-generating level, not only after model training. The prior generates tables that are more structurally aligned with real tabular datasets, particularly in regimes with richer feature interactions. This provides direct evidence that \textsc{O'Prior} compositional realism is doing what it is designed to do: expanding the synthetic task distribution toward realistic dependency structures while preserving broad coverage across schemas and mechanisms.

\subsubsection{Prior Quality via TFM Internal Representations}
\label{sec:internal_representations}

\begin{table}[h!]
\centering
\caption{Accuracy of probing logistic regressors across seven real data and based on all \textsc{nanoTabPFN} layers.}
\textsc{10pt}
\label{tab:tab_probing_all}
\begin{tabular}{llcccccc}
\toprule
Reference Data & Generator & $L1$ & $L2$ & $L3$ & $L4$ & $L5$ & $L6$ \\ 
\midrule
\multirow{4}{*}{Breast Cancer (ID:15)} 
& Full {\sc O'Prior} & 0.98 & 0.98 & 0.98 & 0.96 & 0.96 & 0.96 \\
& \textsc{TabICL}-v2 & 0.98 & 0.98 & 0.98 & 0.98 & 0.97 & 0.97 \\ 
& \textsc{TabPFN} & 0.98 & 0.97 & 0.98 & 0.98 & {\bf 0.99} & {\bf 0.99} \\
& Random (untrained) & 0.98 & 0.98 & 0.98 & 0.98 & 0.98 & 0.98 \\
\midrule

\multirow{4}{*}{Vehicle (ID:54)}
& Full {\sc O'Prior} & 0.33 & {\bf 0.46} & {\bf 0.48} & {\bf 0.58} & {\bf 0.69} & {\bf 0.67} \\
& \textsc{TabICL}-v2 & {\bf 0.39} & 0.38 & 0.39 & 0.39 & 0.38 & 0.40 \\ 
& \textsc{TabPFN} & 0.38 & 0.39 & 0.40 & 0.37 & 0.34 & 0.33 \\ 
& Random (untrained) & 0.33 & 0.37 & 0.39 & 0.38 & 0.39 & 0.39 \\
\midrule

\multirow{4}{*}{Diabetes (ID:37)}
& Full {\sc O'Prior} & 0.70 & {\bf 0.74} & 0.70 & {\bf 0.77} & {\bf 0.76} & {\bf 0.74} \\
& \textsc{TabICL}-v2 & {\bf 0.73} & 0.73 & 0.73 & 0.73 & 0.74 & {\bf 0.74} \\ 
& \textsc{TabPFN} & 0.71 & 0.72 & {\bf 0.74} & 0.72 & 0.71 & 0.66 \\ 
& Random (untrained) & 0.71 & {\bf 0.74} & {\bf 0.74} & 0.74 & 0.73 & 0.73 \\
\midrule

\multirow{4}{*}{Ionosphere (ID:59)}
& Full {\sc O'Prior} & 0.65 & 0.73 & {\bf 0.80} & {\bf 0.80} & {\bf 0.88} & {\bf 0.88} \\
& \textsc{TabICL}-v2 & {\bf 0.73}  & {\bf 0.76} & 0.76 & 0.76 & 0.76 & 0.73 \\
& \textsc{TabPFN} & {\bf 0.73}  & {\bf 0.76} & 0.76 & 0.61 & 0.57 & 0.57 \\ 
& Random (untrained) & 0.65 & 0.69 & 0.69 & 0.69 & 0.69 & 0.69 \\
\midrule

\multirow{4}{*}{Segment (ID:36)}
& Full {\sc O'Prior} & 0.32 & {\bf 0.59} & {\bf 0.68} & {\bf 0.75} & {\bf 0.74} & {\bf 0.83} \\
& \textsc{TabICL}-v2 & 0.27 & 0.55 & 0.56 & 0.56 & 0.53 & 0.53 \\ 
& \textsc{TabPFN} & 0.31 & 0.52 & 0.54 & 0.46 & 0.39 & 0.31 \\ 
& Random (untrained) & {\bf 0.33} & 0.51 & 0.53 & 0.54 & 0.54 & 0.55 \\
\midrule

\multirow{4}{*}{Credit-g (ID:31)}
& Full {\sc O'Prior} & 0.64 & 0.64 & 0.61 & 0.61 & 0.60 & 0.63 \\
& \textsc{TabICL}-v2 & {\bf 0.66} & {\bf 0.66} & 0.64 & 0.64 & {\bf 0.64} & {\bf 0.64} \\
& \textsc{TabPFN} & 0.65 & 0.65 & {\bf 0.65} & {\bf 0.65} & {\bf 0.64} & {\bf 0.64} \\ 
& Random (untrained) & 0.65 & 0.64 & 0.64 & 0.64 & {\bf 0.64} & {\bf 0.64} \\
\midrule

\multirow{4}{*}{Blood (ID:1464)}
& Full {\sc O'Prior} & 0.74 & {\bf 0.78} & {\bf 0.81} & {\bf 0.79} & {\bf 0.79} & {\bf 0.77} \\
& \textsc{TabICL}-v2 & {\bf 0.77} & 0.74 & 0.72 & 0.75 & 0.75 & 0.75 \\ 
& \textsc{TabPFN} & 0.74 & 0.73 & 0.73 & 0.74 & 0.74 & 0.74 \\ 
& Random (untrained) & 0.74 & 0.74 & 0.74 & 0.74 & 0.74 & 0.75 \\
\bottomrule
\end{tabular}
\end{table}

The previous experiments evaluate synthetic priors either through downstream benchmark performance or through structural alignment of the generated tables themselves. We now ask another question: how does the choice of synthetic prior change the internal representations learned by the tabular foundation model? This experiment evaluates whether pretraining on \textsc{O'Prior} induces progressively more informative representations inside the transformer, rather than merely improving final predictions through shallow or dataset-specific effects.

\paragraph{Motivation.} A tabular foundation model trained on a useful prior should not only perform well at the output layer; it should organize real tabular examples into internal representations that become increasingly predictive across depth. If the synthetic pretraining distribution exposes the model to realistic structural mechanisms, heterogeneous feature interactions, missingness patterns, and support--query shifts, then deeper layers should encode more task-relevant information. Conversely, if a prior induces only weak or brittle abstractions, linear probes trained on intermediate representations should remain flat across layers, degrade with depth, or perform no better than an untrained model.

This representation-level evaluation is important because it directly probes what the model has learned from the prior. It separates genuine representation learning from superficial output behavior and provides a mechanistic view of why \textsc{O'Prior} improves transfer. In particular, we expect the benefit of \textsc{O'Prior} to be most visible on datasets with nonlinear structure, multiple classes, and richer feature dependencies, where a model must progressively refine its representation rather than rely on simple marginal or low-level cues.

\paragraph{Evaluation protocol.}
We compare three pretrained synthetic priors, \textsc{O'Prior}, \textsc{TabICL-v2}, and \textsc{TabPFN}, against a Random untrained model. All models share the same \textsc{nanoTabPFN} architecture. For each real reference dataset, we pass the data through the model and extract hidden representations from all six transformer layers, denoted $L1$ through $L6$. At each layer, we freeze the representation and train a logistic regression probe to predict the target labels. The probe accuracy measures how linearly accessible the class information is at that layer.

This protocol evaluates the quality of the representation learned by the prior, not the capacity of the probing model. Since the probe is linear and the architecture is fixed across conditions, differences in probing accuracy reflect how well each synthetic prior shapes the internal geometry of the transformer. The Random untrained baseline is included as a control for architectural and preprocessing effects: a pretrained model is only meaningfully informative when it exceeds the representations produced by random initialization.

\paragraph{Results.} Table~\ref{tab:tab_probing_all} reports probing accuracy across all six layers on seven real datasets. The clearest pattern is that \textsc{O'Prior} produces progressively stronger representations on the datasets where representation learning is genuinely required. On Vehicle, accuracy rises from $0.33$ at $L1$ to $0.69$ at $L5$ and $0.67$ at $L6$, while \textsc{TabICL-v2} remains nearly flat around $0.38$--$0.40$ and \textsc{TabPFN} degrades from $0.38$ to $0.33$. The Random model also remains near the low baseline range. This shows that \textsc{O'Prior} learns class-relevant structure that becomes increasingly accessible with depth, whereas the alternative priors do not produce comparable layerwise refinement.

A similar pattern appears on Ionosphere and Segment. On Ionosphere, \textsc{O'Prior} improves from $0.653$ at $L1$ to $0.884$ at both $L5$ and $L6$, substantially exceeding \textsc{TabICL-v2}, \textsc{TabPFN}, and the Random baseline. On Segment, the trend is even more pronounced: \textsc{O'Prior} increases from $0.32$ at $L1$ to $0.83$ at $L6$, while \textsc{TabICL-v2} reaches only $0.56$ at its best and ends at $0.53$, \textsc{TabPFN} declines to $0.31$, and the Random model remains around $0.55$ at the final layer. These results indicate that \textsc{O'Prior} does not merely encode shallow separability; it builds deeper, progressively more discriminative representations on structurally challenging tabular tasks. 

The results on Blood and Diabetes provide a more moderate but still informative signal. On Blood, \textsc{O'Prior} reaches its strongest probing accuracy at intermediate layers, peaking at $0.81$ in $L3$ and remaining above the competing priors through later layers. On Diabetes, all methods are relatively close, but \textsc{O'Prior} reaches the highest observed score, $0.77$ at $L4$, while \textsc{TabICL-v2} remains nearly constant around $0.73$--$0.74$ and \textsc{TabPFN} decreases to $0.66$ by $L6$. This suggests that even when the task is less demanding or partially linearly separable at early layers, \textsc{O'Prior} still preserves or improves task-relevant information with depth.

Two datasets, Breast Cancer and Credit-g, are less diagnostic. On Breast Cancer, all models, including the Random baseline, achieve very high probing accuracy across layers. This indicates that the dataset is already close to linearly separable under the model input representation, making it unsuitable for distinguishing the effect of synthetic priors. On Credit-g, all methods remain in a narrow range around $0.60$--$0.66$, with no consistent depth-wise improvement. This suggests that either the task is not well captured by the probed representations or that its predictive signal is too weak for this probing setup to differentiate the priors.

\paragraph{Interpretation.} The representation-level evidence supports the central claim of this work: the synthetic prior determines not only downstream accuracy, but also the internal abstractions learned by a tabular foundation model. Across the most informative datasets, \textsc{O'Prior} produces a characteristic layerwise progression: early layers encode weak or local information, while deeper layers become substantially more linearly predictive. This is precisely the behavior expected from a model trained on a richer and more realistic task distribution. The Hybrid SCM component exposes the model to heterogeneous mechanisms and nonlinear dependencies; the realism engine introduces observational artifacts such as irregular marginals, missingness, and target transformations; and the shift-stress modules discourage reliance on brittle shortcuts. Together, these components encourage the transformer to form representations that are robust, structured, and increasingly task-aligned with depth.

By contrast, \textsc{TabICL-v2} often produces flat representations, and \textsc{TabPFN} frequently degrades on the more challenging datasets. This indicates that their pretrained representations are less able to progressively organize real tabular examples into linearly separable structures. The Random baseline is crucial for interpreting this result: on easy datasets such as Breast Cancer, random features already perform well, so high probing accuracy alone is not evidence of useful pretraining. The strongest evidence for \textsc{O'Prior} comes from datasets such as Vehicle, Ionosphere, and Segment, where the Random model and competing priors remain flat or weak, while \textsc{O'Prior} improves substantially across layers.

Overall, this experiment provides mechanistic support for the value of \textsc{O'Prior}. Its gains are not limited to final-task performance; they are reflected inside the model as deeper, more discriminative internal representations. This suggests that compositional realism in the synthetic prior leads to better representation learning, enabling the tabular foundation model to extract and refine class-relevant structure from real datasets more effectively than priors based on simpler or less realistic synthetic task distributions.


\section{Related Work}
\label{sec:discussion-related}

\paragraph{Tabular learning and tabular foundation models.}
Tree-based ensembles, XGBoost \cite{chen2016xgboost}, LightGBM \cite{ke2017lightgbm}, and CatBoost \cite{prokhorenkova2018catboost}, remain the dominant baseline for supervised tabular prediction, owing to inductive biases well-suited to heterogeneous features, irregular marginals, and bounded decision boundaries \cite{grinsztajn2022tree, mcelfresh2023neural}. TFMs have recently emerged as a compelling alternative, achieving competitive performance without dataset-specific training \cite{tabpfn, tabpfnv2.5, tabicl, tabdpt, orionmsp, orionbix, limix}. Our work does not propose a new TFM architecture; it treats the synthetic prior as the primary object of study. Recent work has also developed unified libraries for TFM inference and fine-tuning \cite{tanna2025tabtune} and systematic studies of post-pretraining adaptation \cite{tanna2026finetuning}.

\paragraph{Synthetic task generation for tabular pretraining.}
The prior-data fitted network (PFN) framework \cite{DBLP:conf/iclr/0005HPGH22} introduced the idea of amortizing Bayesian inference over a synthetic prior via a pretrained transformer. \textsc{TabPFN} \cite{tabpfn} instantiated this for tabular classification using SCM-based priors \cite{pearl2010causal,peters2017elements} with random BNN mechanisms, demonstrating competitive performance without task-specific training. Subsequent TFMs extended the paradigm: \textsc{TabPFN v2} \cite{tabpfnv2} enriched the prior alongside architectural changes, making it impossible to attribute gains to the prior alone. \textsc{TabForestPFN} \cite{den2024fine} augmented generation with tree-based decision boundaries; \textsc{TabICL} \cite{tabicl} scaled ICL to 500K samples; \textsc{TabDPT} \cite{tabdpt} departed from synthetic pretraining entirely. Recent work has begun examining SCM-based generation more carefully \cite{DBLP:journals/corr/abs-2603-10254, wang2026relational}, but systematic ablation of prior components remains absent. \textsc{O'Prior} directly addresses this: it is designed to be compositionally ablatable under a controlled protocol that holds architecture, optimization, and compute fixed.

\paragraph{Generative models for tabular data.}
A broad literature targets statistical fidelity to a fixed dataset, spanning GAN-based methods \cite{xu2019modeling}, autoregressive models \cite{great}, and diffusion models \cite{tabddpm}. These methods share a fundamentally different objective from \textsc{O'Prior}: they approximate the distribution of a single observed table rather than defining a distribution over supervised learning problems for general-purpose pretraining.

\section{Conclusion}
\label{sec:conclusion}

This work studies a simple but central question for tabular foundation models: what makes a synthetic pretraining prior useful? Unlike language and vision models, tabular foundation models acquire most of their inductive bias from synthetic pretraining tasks; the prior is therefore not just a data source, but the mechanism that defines what kinds of tabular problems the model learns to solve.

We introduced \textsc{O'Prior}, a compositional realism prior that combines Hybrid SCMs, modular realism perturbations, shift-and-shortcut stress, and a leakage-safe curriculum protocol. Under a controlled evaluation in which architecture, optimizer, compute budget, and evaluation pipeline are fixed, changing only the prior leads to substantial differences in downstream performance. The ablations show that structural mechanism diversity is the strongest driver of transfer, while observational realism and shift-aware stress provide complementary gains.

Beyond benchmark performance, \textsc{O'Prior} also improves the quality of the synthetic task distribution itself. The structural-alignment analysis shows stronger dependency realism relative to real tabular datasets, and the probing experiments show that models pretrained on \textsc{O'Prior} learn deeper, more discriminative internal representations on structurally challenging tasks. Together, these results support the central claim that better priors produce better tabular foundation models not only at the output level, but also in the data-generating distribution and learned representation space.

Overall, \textsc{O'Prior} provides both a practical prior for tabular pretraining and a framework for studying prior construction as a controllable, ablatable design problem. Future work should scale this framework to larger pretraining budgets, evaluate it under targeted distribution-shift benchmarks, and explore domain-specialized realism modules for high-impact tabular applications.

\bibliographystyle{unsrt}
\bibliography{references}

\newpage
\appendix
\section{Data Visualisation}
\label{app:viz}
\begin{figure}[t]
    \centering
    \includegraphics[width=\linewidth]{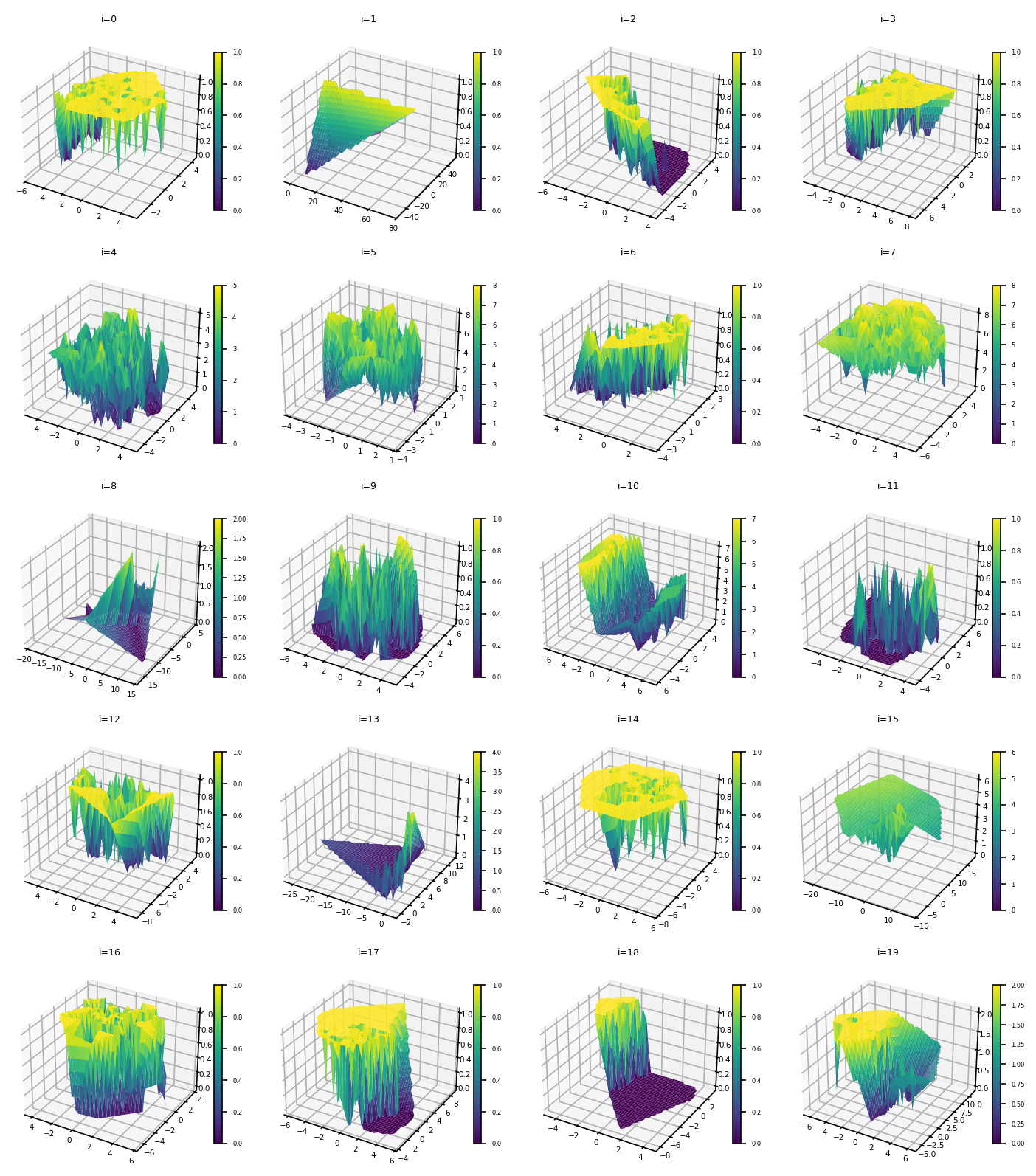}
    \caption{Distribution of data generated based on G1c (SM+SH).}
    \label{fig:variant_1d}
\end{figure}

\begin{figure}[t]
    \centering
    \includegraphics[width=\linewidth]{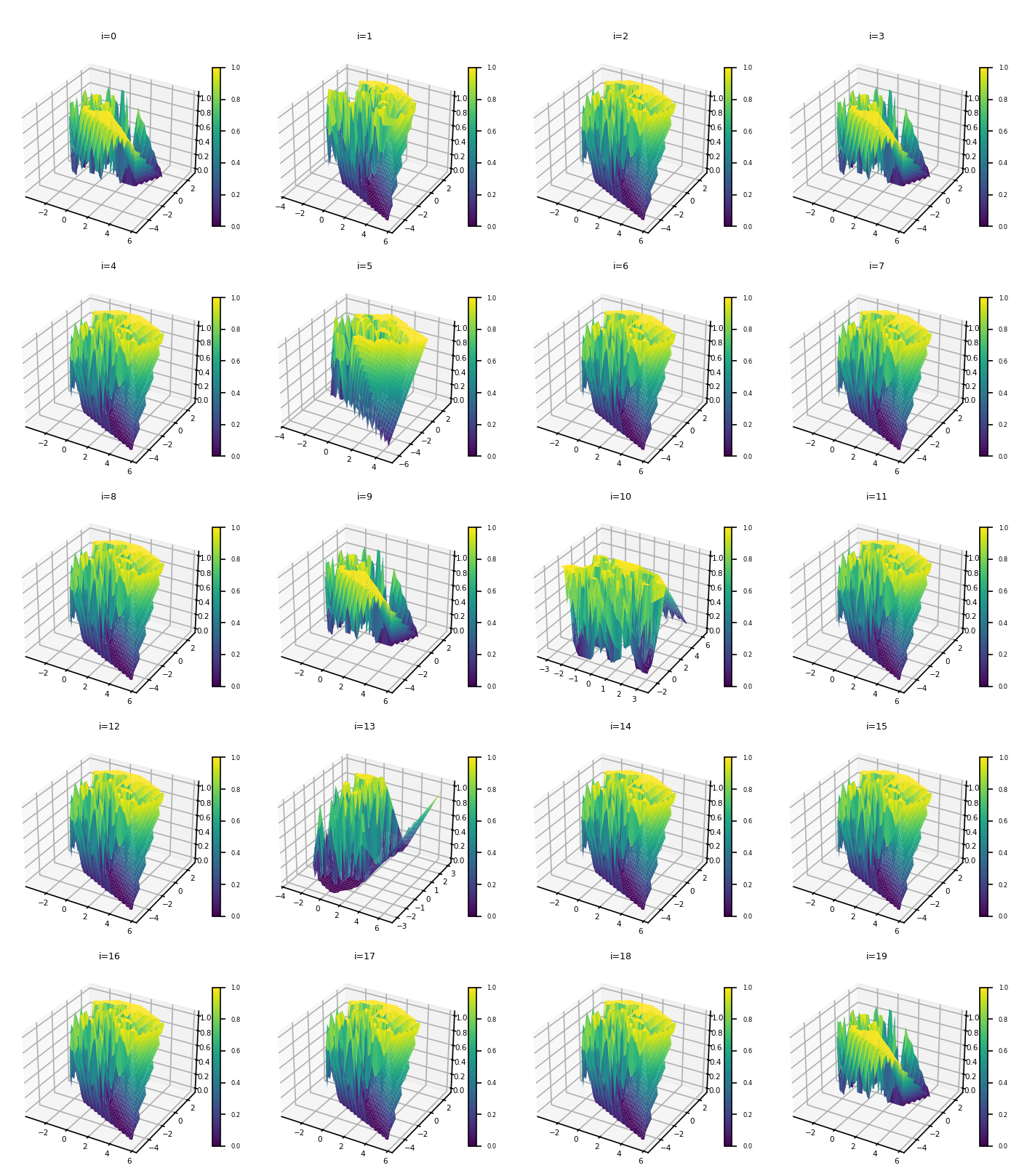}
    \caption{Distribution of data generated based on G2c (SM+SH+SR).}
    \label{fig:variant_2c}
\end{figure}

\begin{figure}[t]
    \centering
    \includegraphics[width=\linewidth]{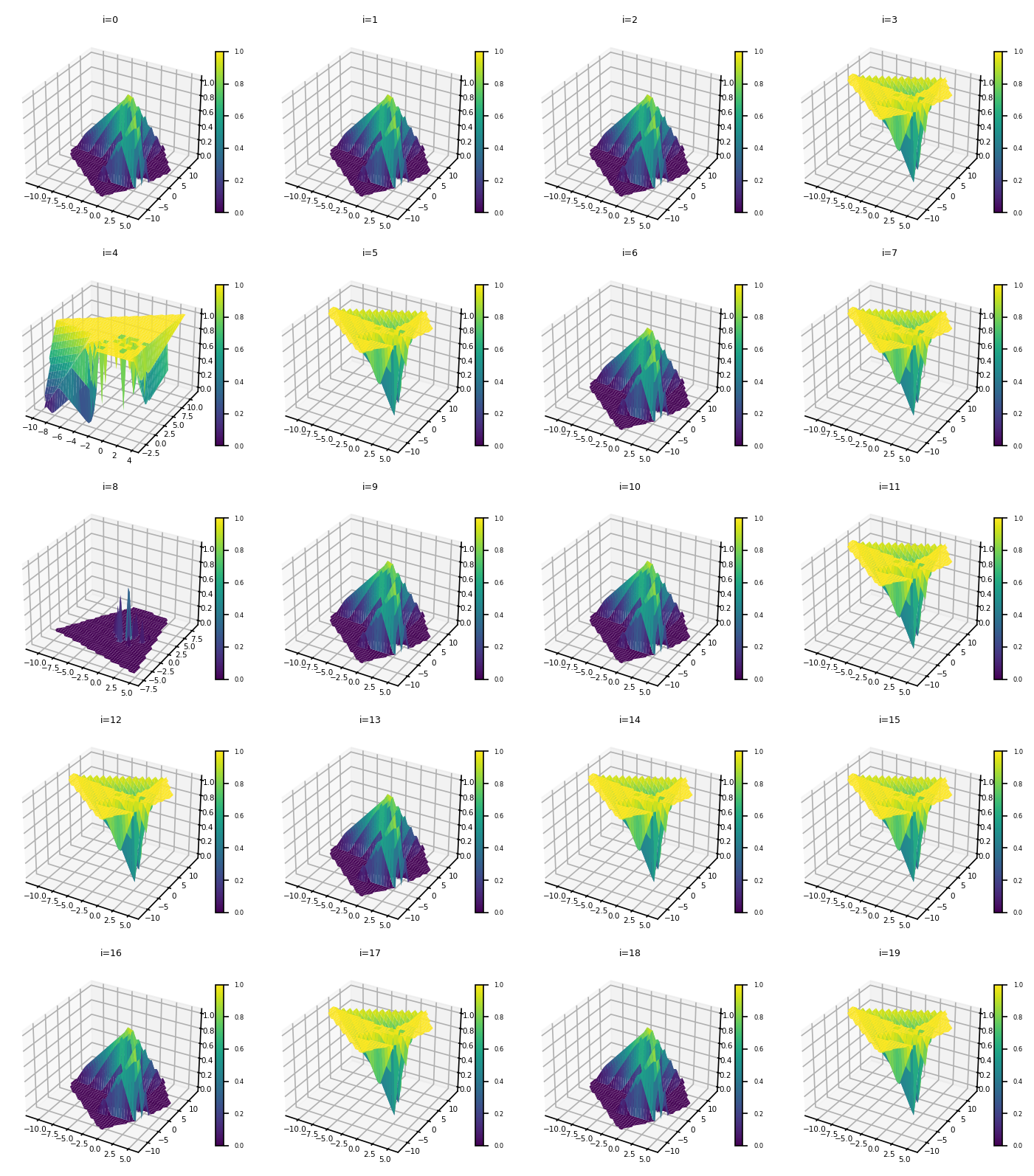}
    \caption{Distribution of data generated based on G3b (SM+SH+SD).}
    \label{fig:variant_3b}
\end{figure}

\begin{figure}[t]
    \centering
    \includegraphics[width=\linewidth]{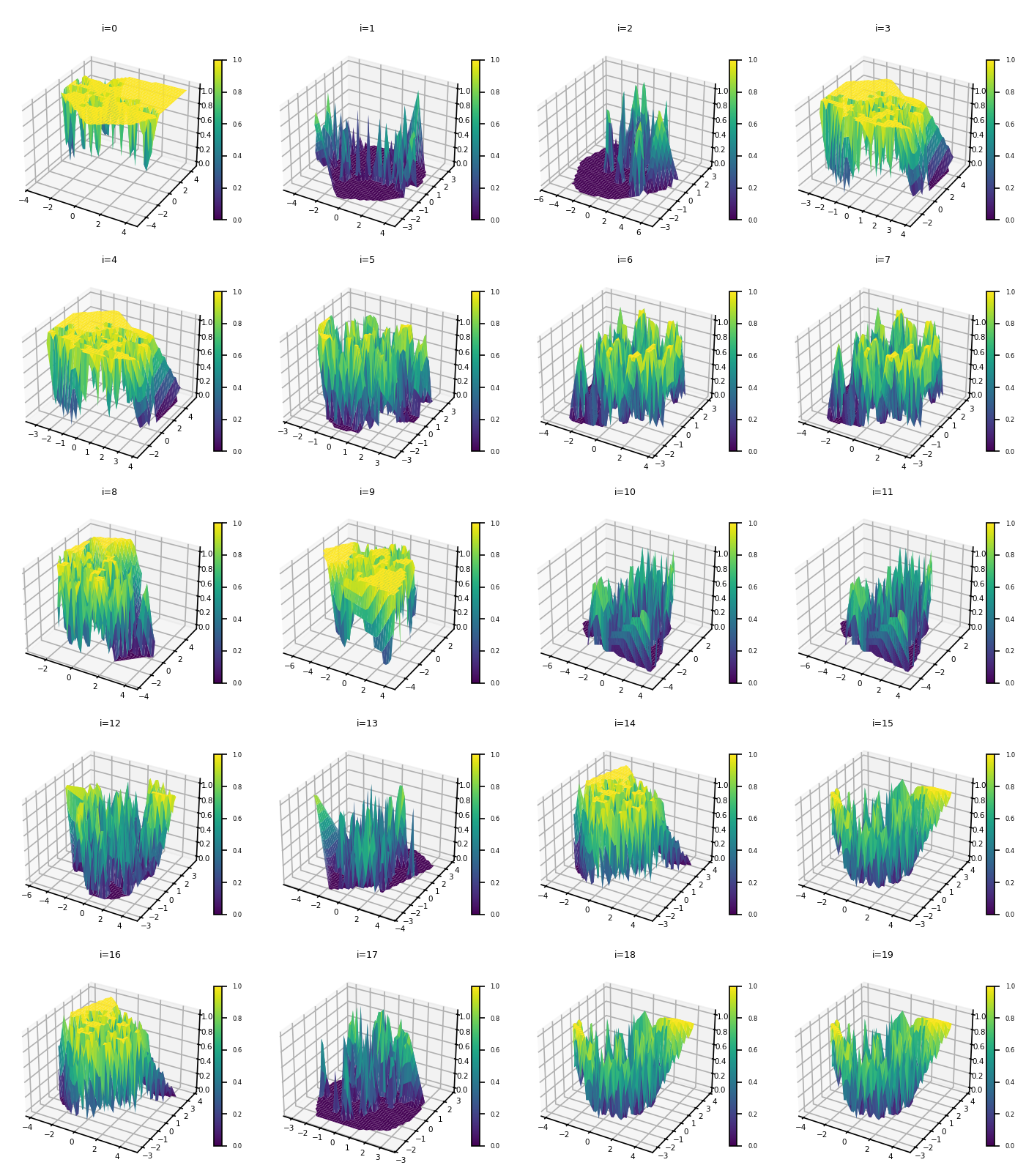}
    \caption{Distribution of data generated based on G4 (G3b with Curriculum).}
    \label{fig:variant_4}
\end{figure}

Figures~\ref{fig:variant_1d},~\ref{fig:variant_2c},~\ref{fig:variant_3b},~\ref{fig:variant_4} shows the PCA-based distributions of G1c, G2c, G3b, G4 variants respectively, introduced in Section~\ref{sec:protocol} and Table~\ref{tab:ablation_variants}. We observe that G1c is the most unpredictable where scale can be different across datasets (i.e., $i=1$ x-axes reaching $80$ units), where the other are more bounded. This suggests the hybrid mechanism may amplify variance with no stabilising normalisation. Oh the other head, variant G2c is the most controlled as it tends to produce the same shape (i.e., a high plateau fading toward one edge) across multiple runs making this variant the most predictable for downstream use. This is due to the injection of statistical properties like skewness and missingness which results in distributions that are well-bounded. G3b variants tend to produce data that is organised within a pyramid or tent shape in one region, leaving similar-valued areas. This is due to the use of multiple distributional shifts (i.e., seasonal, covariate, and latent confounding) which forbid the spreading of data evenly across the whole space. Finally, G4 combines all properties of previous while addresses G3b's sparsity problem as distributions look denser in Figure~\ref{fig:variant_4} than Figure~\ref{fig:variant_3b}. We also observe some high plateau shapes as in Figure~\ref{fig:variant_2c}, and diversity of surfaces as in Figure~\ref{fig:variant_1d}, which making this variant the most diverse of all variants in terms of the breadth of distributional scenarios it covers.

\section{Benchmark Dataset Details}
\label{app:datasets}

This appendix reports the datasets used in the experimental evaluation. The table includes datasets from the TabArena benchmark \ref{tab:tabarena-datasets} and the OpenML-CC18 benchmark\ref{tab:openml-datasets}. For each dataset, we provide the benchmark source, the resolved OpenML dataset identifier, dataset name, number of samples, number of features, number of classes, and the corresponding task type. 

\begingroup
\small
\setlength{\tabcolsep}{4pt}
\renewcommand{\arraystretch}{0.95}

\begin{longtable}{p{1.5cm}p{4.8cm}r r c p{2.8cm}}
\caption{Summary of the selected TabArena-v0.1 classification datasets used in the evaluation.}
\label{tab:tabarena-datasets}\\

\toprule
Dataset ID & Dataset name & Samples & Features & Classes & Task type \\
\midrule
\endfirsthead

\toprule
Dataset ID & Dataset name & Samples & Features & Classes & Task type \\
\midrule
\endhead

\midrule
\multicolumn{6}{r}{Continued on next page}\\
\endfoot

\bottomrule
\endlastfoot

46941 & maternal\_health\_risk & 1,014 & 7 & 3 & multiclass classification \\
46905 & Amazon\_employee\_access & 32,769 & 10 & 2 & binary classification \\
46906 & anneal & 898 & 39 & 5 & multiclass classification \\
46952 & qsar-biodeg & 1,054 & 42 & 2 & binary classification \\
46908 & APSFailure & 76,000 & 171 & 2 & binary classification \\
46910 & bank-marketing & 45,211 & 14 & 2 & binary classification \\
46911 & Bank\_Customer\_Churn & 10,000 & 11 & 2 & binary classification \\
46912 & Bioresponse & 3,751 & 1,777 & 2 & binary classification \\
46913 & blood-transfusion-service-center & 748 & 5 & 2 & binary classification \\
46915 & churn & 5,000 & 20 & 2 & binary classification \\
46916 & coil2000\_insurance\_policies & 9,822 & 86 & 2 & binary classification \\
46963 & website\_phishing & 1,353 & 10 & 3 & multiclass classification \\
46918 & credit-g & 1,000 & 21 & 2 & binary classification \\
46919 & credit\_card\_clients\_default & 30,000 & 24 & 2 & binary classification \\
46920 & customer\_satisfaction\_in\_airline & 129,880 & 22 & 2 & binary classification \\
46921 & diabetes & 768 & 9 & 2 & binary classification \\
46922 & Diabetes130US & 71,518 & 48 & 2 & binary classification \\
46980 & MIC & 1,699 & 112 & 8 & multiclass classification \\
46924 & E-CommereShippingData & 10,999 & 11 & 2 & binary classification \\
46927 & Fitness\_Club & 1,500 & 7 & 2 & binary classification \\
46938 & Is-this-a-good-customer & 1,723 & 14 & 2 & binary classification \\

\end{longtable}

\endgroup

\begingroup
\small
\setlength{\tabcolsep}{4pt}
\renewcommand{\arraystretch}{0.95}

\begin{longtable}{p{1.5cm}p{4.8cm}r r c p{2.8cm}}
\caption{Summary of OpenML-CC18 datasets used in the evaluation.}
\label{tab:openml-datasets}\\

\toprule
Dataset ID & Dataset name & Samples & Features & Classes & Task type \\
\midrule
\endfirsthead

\toprule
Dataset ID & Dataset name & Samples & Features & Classes & Task type \\
\midrule
\endhead

\midrule
\multicolumn{6}{r}{Continued on next page}\\
\endfoot

\bottomrule
\endlastfoot

11 & balance-scale & 625 & 5 & 3 & multiclass classification \\
15 & breast-w & 699 & 10 & 2 & binary classification \\
18 & mfeat-morphological & 2,000 & 7 & 10 & multiclass classification \\
23 & cmc & 1,473 & 10 & 3 & multiclass classification \\
29 & credit-approval & 690 & 16 & 2 & binary classification \\
31 & credit-g & 1,000 & 21 & 2 & binary classification \\
37 & diabetes & 768 & 9 & 2 & binary classification \\
50 & tic-tac-toe & 958 & 10 & 2 & binary classification \\
54 & vehicle & 846 & 19 & 4 & multiclass classification \\
188 & eucalyptus & 736 & 20 & 5 & multiclass classification \\
307 & vowel & 990 & 13 & 11 & multiclass classification \\
469 & analcatdata\_dmft & 797 & 5 & 6 & multiclass classification \\
1063 & kc2 & 522 & 22 & 2 & binary classification \\
1067 & kc1 & 2,109 & 22 & 2 & binary classification \\
1068 & pc1 & 1,109 & 22 & 2 & binary classification \\
1462 & banknote-authentication & 1,372 & 5 & 2 & binary classification \\
1464 & blood-transfusion-service-center & 748 & 5 & 2 & binary classification \\
1480 & ilpd & 583 & 11 & 2 & binary classification \\
1489 & phoneme & 5,404 & 6 & 2 & binary classification \\
1497 & wall-robot-navigation & 5,456 & 25 & 4 & multiclass classification \\
23381 & dresses-sales & 500 & 13 & 2 & binary classification \\
40701 & churn & 5,000 & 21 & 2 & binary classification \\
40975 & car & 1,728 & 7 & 4 & multiclass classification \\
40982 & steel-plates-fault & 1,941 & 28 & 7 & multiclass classification \\
40983 & wilt & 4,839 & 6 & 2 & binary classification \\
40984 & segment & 2,310 & 20 & 7 & multiclass classification \\
40994 & climate-model-simulation-crashes & 540 & 21 & 2 & binary classification \\

\end{longtable}

\endgroup

\end{document}